\pdfoutput=1

\documentclass[11pt]{article}

\usepackage[final]{acl}

\usepackage{times}
\usepackage{latexsym}
\usepackage{tabularray}
\usepackage{xcolor}
\definecolor{baselinegrey}{RGB}{240,240,240}
\definecolor{mygrey}{RGB}{230,230,230}
\usepackage{soul}
\usepackage{todonotes}
\usepackage{enumitem}
\usepackage{lingmacros}

\usepackage[T1]{fontenc}

\usepackage[utf8]{inputenc}

\usepackage{microtype}

\usepackage{inconsolata}

\title{
From \textit{Showgirls} to \textit{Performers}: 
Fine-tuning with Gender-inclusive Language for Bias Reduction in LLMs}

\author{Marion Bartl \and Susan Leavy \\
        Insight SFI Research Centre for Data Analytics \\
        School of Information and Communication Studies\\
        University College Dublin\\
        \texttt{marion.bartl@insight-centre.org}\\
        \texttt{susan.leavy@ucd.ie}}

\begin{document}
\maketitle
\begin{abstract}
Gender bias is not only prevalent in Large Language Models (LLMs) and their training data, but also firmly ingrained into the structural aspects of language itself. 
Therefore, adapting linguistic structures within LLM training data to promote gender-inclusivity can make gender representations within the model more inclusive.
The focus of our work are gender-exclusive affixes in English, such as in \textit{show\underline{girl}} or \textit{\underline{man-}cave}, which can perpetuate gender stereotypes and binary conceptions of gender.
We use an LLM training dataset to compile a catalogue of 692 gender-exclusive terms along with gender-neutral variants and from this, develop a gender-inclusive fine-tuning dataset, the \textit{Tiny Heap}. 
Fine-tuning three different LLMs with this dataset, we observe an overall reduction in gender-stereotyping tendencies across the models. Our approach provides a practical method for enhancing gender inclusivity in LLM training data and contributes to incorporating queer-feminist linguistic activism in bias mitigation research in NLP. 

\end{abstract}

\section{Introduction}

Large language models (LLMs) have become ubiquitous in Natural Language Processing (NLP) due to their impressive capabilities in a variety of tasks. However, they also carry risks arising from social biases incorporated into models from the training data~\citep{bender_dangers_2021}. Well-documented among these are harmful gender biases such as 
reliance on stereotypes and erasure of non-binary gender identities~\citep[][a.o.]{cao_toward_2021, ovalle_im_2023}. 
Structural aspects of language itself and linguistic norms can reflect as well as shape societal concepts of gender~\citep{pauwels_linguistic_2003, whorf_language_1956}. Within the context of LLMs, encoded representations of gender inform language generation and classification decisions, thereby having the potential to influence societal concepts of gender~\citep{bommasani_opportunities_2022}. It is vital therefore, to ensure that LLMs are evaluated and trained to minimize gender bias and promote equitable representation of all genders. 

In English, linguistic structures have a long history of reinforcing traditional gender roles and the concept of male gender as the default~\citep{mills_gender_2012}. Examples include the use of \textit{man} to mean all humans, the indication of women's marital status in terms of address (\textit{Miss}, \textit{Mrs.}, \textit{Ms.}), or the marking of deviation from gendered norms (\textit{male nurse}, \textit{girl boss}). Sexist and gender-exclusive linguistic constructions have been discouraged in official style guides~\citep{apa_publication_2020} and their use has been in decline~\citep{baker_will_2010}. However, the nature of language change is slow, with new and traditional variations existing simultaneously. Given the scale of LLM training data~\citep{bender_dangers_2021} and the disproportionate representation of men within textual data~\cite{baker_sociolinguistics_2010}, language models have the potential to proliferate and reinforce stereotypical and traditional views of gender.

Approaches to mitigating bias in LLMs have included fine-tuning with \textit{gender-inclusive language}~\citep{thakur-etal-2023-language}. Data interventions with gender-inclusive text aim to reduce the use of binary gender terms in cases where gender is irrelevant (for example, a \textit{chairman} and \textit{chairwoman} do the same job) and thereby allow for association of a term with all genders (\textit{chairperson}).
However, the replacement of sexist and gender-exclusive terminology often relies on limited lists of gender-neutral terms~\citep{ghanbarzadeh-etal-2023-gender,thakur-etal-2023-language}, and often focuses on professions~\citep{fatemi-etal-2023-improving}. Additionally, previous works on fine-tuning LLMs with gender-inclusive data have primarily carried out experiments with masked language models such as BERT~\citep{devlin_bert_2019} and its derivatives~\citep{vashishtha-etal-2023-evaluating}. 

In this research, we focused on expanding the coverage of gender-exclusive terminology and experimented with fine-tuning both causal and masked LLMs. 
We first exploited structural elements of English that relate to gender discrimination and exclusion in order to generate a larger catalogue of words that are unnecessarily gendered along with gender neutral alternatives.
We extracted nouns with gender-marking prefixes and suffixes from a common training corpus, OpenWebText2~\citep{gao_pile_2020}, which was used to train LLMs like Meta's Llama2~\citep{thakur-etal-2023-language} and Microsoft's MT-NLG~\citep{smith_using_2022}. The distribution of extracted gender-marking nouns demonstrated clear androcentric tendencies within the corpus.
We compiled gender-neutral variants for each term with a gender-marking affix to form a catalogue of 692 term pairs. This resource is just over three times larger than the size of previously available resources and could be used in assessments of gender skew within LLM training corpora as well as in the replacement gender-exclusive terminology. 
We also developed a small-scale, multi-domain fine-tuning corpus, using our catalogue to replace gender-exclusive with gender-neutral words. We also employed the NeuTral Rewriter~\citep{vanmassenhove_neutral_2021} to replace gendered pronouns (\textit{he}, \textit{she}, \textit{himself} etc.) with singular \textit{they}. The resulting corpus was used to fine-tune three different (masked and causal) LLMs. The results of this process of fine-tuning with gender inclusive terminology demonstrated an overall tendency towards reduction in gender-stereotyping exhibited by the models as well as a reduction in the generation of harmful language in gendered contexts.

\paragraph{Contributions}
\begin{itemize}[noitemsep]
    \item We show clear androcentric tendencies within a commonly used LLM training corpus.
    \item We construct a catalogue of 692 term pairs, consisting of a gender-exclusive terms and neutral alternatives, which we release for public use\footnote{\url{https://github.com/marionbartl/affixed_words}}.
    \item We show that automatically generated gender-inclusive English is effective in reducing gender stereotyping in LLMs through fine-tuning\footnote{\url{https://github.com/marionbartl/performers}}.
\end{itemize}

\section{Bias Statement}
The focus of this work is gender-inclusive language, and its counterpart, sexist language. Sexist language, following~\citeposs{frye_sexism_1983} definition of sexism, can be defined as language that clearly divides between two genders, in which one gender (masculine) is treated as hierarchically superior to the other (feminine). This superiority is expressed, for example, through the generic use of masculine gendered expressions (e.g. use of terms such as \textit{mankind}, \textit{chairman} to refer to people of any gender). 

Our work is based on the assumption that sexist language in training data is one of the sources of gender bias in LLMs. Specifically, we would expect models to favor masculine expressions over gender-neutral alternatives, creating a representational harm for people of non-masculine gender~\citep{blodgett_language_2020}. Sexist expressions additionally reinforce traditional gender roles (e.g. \textit{male nurse}), therefore we would also expect models to favor gender-stereotypical expressions.
Moreover, since sexist language is based on a binary model of gender, we expect models to default to this. This can lead to misrepresentation and erasure of non-binary genders in downstream applications, creating allocational and representational harms for non-binary users of these systems~\citep{blodgett_language_2020}.
Not adjusting LLMs to accurately represent the variety of genders that exist in society will contribute to the ongoing marginalization of people identifying as gender-queer~\citep{ovalle_im_2023}.

\section{Related Work}
Large Language Models (LLMs) have been shown to encode a variety of social biases contained in their training data~\citep{gupta_survey_2023, salinas_unequal_2023}, among them gender bias~\citep{stanczak_survey_2021}. 
Due to the current prevalence of transfer learning in NLP, in which a pre-trained model is fine-tuned with task-specific data, transfer learning has recently also been adapted by works that aimed to reduce gender bias in LLMs~\citep{lauscher_sustainable_2021, ghanbarzadeh-etal-2023-gender}. In this approach, an LLM is fine-tuned with data that has undergone interventions to increase gender fairness. 
This approach is supported by the finding that biases in fine-tuning data have a greater influence on downstream model behavior than biases in the pre-training data~\citep{steed_upstream_2022}. 
Previous interventions to fine-tuning data include Counterfactual Data Augmentation (CDA), in which masculine and feminine pronouns and gendered nouns are swapped for the respective other~\citep{ghanbarzadeh-etal-2023-gender, vashishtha-etal-2023-evaluating, fatemi-etal-2023-improving}. 
Another intervention replaces gendered words for gender-neutral words (\textit{fire fighter} for \textit{fireman}) or phrases containing both masculine and feminine genders (\textit{he and she} for \textit{he};~\citeauthor{thakur-etal-2023-language},~\citeyear{thakur-etal-2023-language}). 
This kind of intervention is not new: it rests upon a longstanding tradition of research and advocacy the field of feminist linguistics, which has been promoting changes in the lexicon to reduce gender stereotyping and masculine-default language since the 1970s~\citep{kramer_feminist_2016,mills_gender_2012,lakoff_language_1973}. 
More recently such changes to the language, also called \textit{feminist language reform}, have incorporated ways of adapting language to include non-binary and trans gender identities, such as the third person singular (neo)pronouns (\textit{they}, \textit{xe}, \textit{ze}, etc.). The usage and possible modelling of this extended lexicon of pronouns within the context of NLP was analyzed by~\citet{lauscher_welcome_2022}.
\citet{lund-etal-2023-gender} also showed that training on data containing singular \textit{they} can reduce gender bias in grammatical error correction. 
Furthermore,~\citet{vanmassenhove_neutral_2021} and~\citet{sun_they_2021} developed rule-based and neutral machine translation-based models to modify English text to render it gender-neutral.~\citeposs{vanmassenhove_neutral_2021} NeuTral Rewriter replaces gendered pronouns with singular \textit{they} and a list of gendered nouns with neutral variants.
However, while the amount of NLP research incorporating and exploring strategies of feminist language reform has grown, the queer-feminist linguistic research it is based on is, with some exceptions~\citep{devinney_theories_2022, piergentili-etal-2023-gender, seaborn_transcending_2023}, rarely acknowledged and even less often informs the research itself. 

\section{Method}
Gender bias in the English language is reflected in features such as masculine generics and is captured in datasets through, for example, skewed distributions of pronouns and profession words in the same context. However, it is also contained in structural elements of the language itself, such as gender-marking affixes. The most frequent are suffixes such as \textit{-man} in \textit{spokes\underline{man}}, but gender can also be marked with a prefix, such as in \textit{\underline{man-}power} or \textit{\underline{girl}boss}. 
Words marked with masculine suffixes have traditionally been used in a generic sense (e.g. \textit{Madam Chairman}), however, with the emergence of feminist language reform, style guides have advised against their use~\citep{piergentili_gender_2023}. In English, the most common replacement strategy for gendered generics is neutralisation (\textit{chairperson}), because all gender identities, not just male and female, can be referred to by gender-neutral nouns. 
In NLP, research using gender-neutral language in the context of English LLMs has mainly relied on lists of common gender-neutral replacements~\citep{vanmassenhove_neutral_2021, thakur-etal-2023-language}, without taking structural processes such as affixation into account in order to broaden the coverage of these lists.

In this section we first outline the process of extracting unnecessarily gendered words based on gender-marking affixes (§\ref{ssec:extraction}). We then describe the gender-neutralizing interventions to our fine-tuning data (§\ref{ssec:fine-tuning}) as well as the models (§\ref{ssec:models}) and bias measurements used (§\ref{ssec:bias_measure}). 

\subsection{Word Catalogue}\label{ssec:extraction}

\begin{table}
\centering
\begin{tblr}{
    colspec={X[1.1,c]X[2,c]X[c]X[c]X[c]},
    row{1,6,13,14,15} = {font=\bfseries},
    row{1} = {m},
    column{1} = {font=\bfseries},
    stretch = 0.5,
    }
    \hline
    \SetCell[c=2]{c,m}affix & & round 1 & round 2 & round 3 \\ \hline
    \SetCell[r=5]{c} prefix & \textit{woman-} & 10 & 4 & 4 \\
    & \textit{girl-} & 30 & 13 & 10 \\ 
    & \textit{man-} & 87 & 47 & 49 \\ 
    & \textit{boy-} & 59 & 11 & 7 \\ \hline
    & total & 186 & 75 & 70 \\ \hline
    \SetCell[r=6]{c} suffix & \textit{-woman} & 42 & 37 & 35 \\
    & \textit{-girl} & 47 & 24 & 14 \\ 
    & \textit{-man} & 271 & 238 & 180 \\
    & \textit{-boy} & 62 & 41 & 24 \\ \hline
    & \textit{-womanship} & 2 & 2 & 2 \\
    & \textit{-manship} & 53 & 32 & 30 \\ \hline
    & total & 477 & 342 & 285 \\ \hline
    \SetCell[c=2]{c} TOTAL & & 663 & 417 & 355 \\ \hline
    \SetCell[c=2]{c} PERCENT & & 100\% & 62.9\% & 53.54\% \\ \hline
\end{tblr}
\caption{Number of singular nouns with gender-marking affixes extracted from subsection of OpenWebText2 corpus throughout verification process.}
\label{tab:extraction_rounds}
\end{table}

\begin{table*}[]
    \begin{tblr}{
    colspec={X[c]X[0.3,c]X[c]X[0.3,c]X[c]X[0.3,c]X[c]X[0.3,c]},
    row{1} = {font=\bfseries},
    stretch = 0,
    }
    \hline
    \textit{-man} & \# & \textit{-woman} & \# & \textit{-boy} & \# & \textit{-girl} & \#\\ \hline
    spokesman & 44,004 & spokeswoman & 14,044 & cowboy & 1167 & showgirl & 46 \\
    congressman & 4,551 & congresswoman & 419 & fanboy & 388 & fangirl & 42 \\
    businessman & 3,830 & businesswoman & 231 & playboy & 374 & cowgirl & 39 \\
    policeman & 3,015 & policewoman & 151 & tomboy & 199 & playgirl & 6 \\
    freshman & 1,055 & anchorwoman & 40 & busboy & 71 & babygirl & 4 \\
    fisherman & 991 & forewoman & 33 & paperboy & 69 & ballgirl & 4 \\
    cameraman & 910 & everywoman & 30 & homeboy & 47 & camgirl & 4 \\
    statesman & 671 & noblewoman & 21 & plowboy & 32 & papergirl & 4 \\
    defenseman & 571 & spokewoman & 19 & bellboy & 16 & tomgirl & 3 \\
    madman & 505 & charwoman & 16 & callboy & 13 & schoolgirl & 3 \\
    \hline
    \end{tblr}
    \caption{Top 10 words with gender-denoting suffixes after second round of verification and their frequencies within 200-million token subset of OpenWebText2}
    \label{tab:top_10_suffixes}
\end{table*}

We extracted words with the suffixes \textit{-man}, {\textit{-manship}}, \textit{-woman}, \textit{-womanship}, \textit{-boy}, \textit{-girl} and words with the prefixes \textit{man-}\footnote{Words with \textit{man-} prefixes were only included if they also had the dash (-) following \textit{man}, because otherwise the false positive rate (\textit{manager}, \textit{mandate}, etc.) would have been too high.}, \textit{woman-}, \textit{boy-} and \textit{girl-}. 
We used a 200 million token random subsection of the OpenWebText2 corpus~\citep{gao_pile_2020} for extraction. 
The words were extracted using regular expressions within Python. We additionally filtered the words to include only English singular nouns. We only filtered for singular nouns to reduce the amount of redundant extractions, and to simplify the dictionary verification later on. Plurals for all verified words were added after the third round of verification.

The \textbf{first round} of verification of extracted affixed terms generally followed a human-in-the-loop approach, meaning that after 20 files, each 1MB in size, the extracted words were manually checked for validity. This eliminated a variety of false positives such as words in which affixes did not denote gender (\textit{german}, \textit{ramen}), spelling errors (\textit{camerman}, \textit{sopkesman}), surnames (\textit{zimmerman}), and other word creations (\textit{heythereman}, \textit{mrfredman}). 
In total, 663 words were extracted in the first round (ref. Table \ref{tab:extraction_rounds}).

After extraction, the terms were verified in the \textbf{second round} using the API of the BabelNet encyclopedic dictionary~\citep{navigli_babelnet_2012}. BabelNet was chosen due to its broad coverage of lexical resources; its search engine combines entries from WordNet, Wikidata and Wikipedia among others. Terms that did not return an entry in BabelNet were disregarded in order to eliminate less established terms, slang and sexually charged terminology. 
If a term contained a dash, such as in \textit{man-bun}, but could not be found in BabelNet, we also searched for the term with a space instead of the dash to not disregard terms due to spelling differences. 
Table \ref{tab:top_10_suffixes} shows the top ten words containing the four simple gender-marking suffixes and their frequency. The highest frequent words with gendered prefixes, and words with \textit{-wo/manship} suffixes are shown in Table \ref{tab:top_10_prefixes} and \ref{tab:manship_womanship} in the Appendix, respectively.

Following the BabelNet verification, words were manually filtered in the \textbf{third round} to exclude words not related to gender (e.g. \textit{boycott}, \textit{boyne}), and proper names such as surnames or words related to pop culture (\textit{batgirl}, \textit{rainman}). 
Furthermore, terms that occurred with a feminine suffix (\textit{noblewoman}) but did not have a masculine equivalent (\textit{nobleman}) were added as their masculine variant to the list, because we treat gender-marking suffixes as exchangeable to mark a different gender. 
The third round left 353 singular affixed nouns.


\subsubsection{Gender-neutral variants}
Gender-neutral variants were manually compiled for all extracted words with gender-marking affixes. A single variant was added for all items in the list to simplify the replacement process. The final gender-neutral variants were discussed and agreed upon by the researchers. 
The proposed replacements are not intended to be definitive substitutes for their gender-marked counterparts. Instead, they were developed for the present experiments to provide gender-neutral terms, as no official list exists. 

\paragraph{Suffixes}
Some gender-marking suffix could simply be exchanged for one that is gender neutral, such as in the common neutralisation of \textit{chair-man/-woman} to \textit{chairperson}. However, this simple replacement does not always work. For example, some frequent terms already have gender-neutral replacements such as \textit{fire fighter} for \textit{fireman} or \textit{police officer} for \textit{policeman}. In these cases, \textit{*fireperson} or \textit{*policeperson} would be ungrammatical\footnote{As per linguistic convention we mark ungrammatical terms with a leading asterisk (*).}. A similar case can be made for less frequent words for which more elegant solutions are available than simply replacing \textit{-man}/\textit{-woman} with \textit{-person}. One approach is to find more fitting suffixes or compound nouns, such as in the neutralisation of \textit{crewman} with \textit{crew member}. Another approach is to replace a word with a gender-neutral synonym, such as in the replacement of \textit{hitman} with \textit{assassin}. A third approach applies to words containing a verb as their root, such as the word \textit{huntsman}, which has the root \textit{hunt}. Here, the word can be replaced by a nominalization: \textit{hunter}.

\paragraph{Prefixes} 
In the case of words with gender-marking prefixes, gender-neutral variants can be constructed by removing the prefix. For example, the word \textit{man-crush} can be neutralised to \textit{crush}. 

Once the list of singular word pairs was fixed, the plural version of every word-pair was added to the final list. The plurals were obtained using the \texttt{inflect} library in Python (version 7.0.0). After adding plurals, we performed one last round of manual verification to ensure all plurals were formed correctly. The final list contains 692 term pairs. For comparison,~\citet{vanmassenhove_neutral_2021} used a list of 91 term pairs. A sample of our final list can be found in Table \ref{tab:catalogue} in the Appendix.

\subsection{Fine-Tuning Data}\label{ssec:fine-tuning}

\begin{table}[h]
    \centering
    \begin{tblr}{
    colspec={X[1.5, l]X[c]X[c]X[c]X},
    hline{2,3,6,7} = {solid},
    row{1,2} = {font=\bfseries, m, c},
    rows = {m},
    stretch = 0.2,
    }
    & & Heap & Small Heap & Tiny Heap\\
    dataset & original weight & \SetCell[c=3]{c,m} \# tokens & & \\
    OWT2 & 50\%  & 125M & 25M & 162k\\
    CC-News & 30\% & 75M & 15M & 240k \\
    English Wikipedia & 20\%  & 50M & 10M & 112k \\
    TOTAL & 100\% & 250M & 50M & 514k \\
    \end{tblr}
    \caption{Composition of Heap corpora; OWT2 = OpenWebText2, CC-News = Common Crawl News} 
    \label{tab:corpus_composition}
\end{table}

\begin{table*}[t]
\centering
\begin{tblr}{
    colspec={X[0.3,l]X},
    column{1} = {font=\bfseries},
    stretch = 0.7,
    rows = {m},
    hlines,
} 
original sentence & He told \underline{\textit{newsmen}} at the scene that unknown criminals vandalised MD metres and armoured cables of the transformer. \\
after word replacement & He told \underline{\textit{reporters}} at the scene that unknown criminals vandalised MD metres and armoured cables of the transformer. \\
after rewriting and word replacement & \underline{\textit{They}} told \underline{\textit{reporters}} at the scene that unknown criminals vandalised MD metres and armoured cables of the transformer.\\
\end{tblr}
\caption{\label{tab:rewriting_example}
Example of sentences in fine-tuning data at different stages of gender-neutral rewriting and replacement
}
\end{table*}

To create a fine-tuning corpus with gender-neutral interventions, we assembled a base corpus, which needed to have several features:  
(1) The configuration should be similar to current LLM pre-training data, meaning that it should contain a diverse set of sources. However, we excluded data that was too domain-specific, such as code and scientific publications in order to demonstrate methodology for general-purpose English. In the same line of reasoning,
(2) the corpus should only contain English data, because the focus of this work is English, and 
the NeuTral Rewriter~\citep{vanmassenhove_neutral_2021}, which replaces gendered pronouns with singular \textit{they} does also only exist for English.
(3) Finally, since we do not aim to worsen the performance of the LLM through fine-tuning, the corpus should only include high-quality text. 

The final composition of our base corpus was inspired by the composition of GPT-3's training data~\citep{brown_language_2020} as well as The Pile corpus~\citep{gao_pile_2020} and is shown in Table \ref{tab:corpus_composition}. Our original download has a size of 250 million tokens, which is approximately 1.5 GB of data. Since this is substantially smaller than The Pile (825GB), we called our dataset \textit{\textbf{The Heap}}. The dataset was downloaded using the Huggingface \texttt{datasets} library (version 1.18.3;~\citeauthor{wolf_transformers_2020},~\citeyear{wolf_transformers_2020}) and tokenized with the \texttt{stanza} library (version 1.7.0;~\citeauthor{qi_stanza_2020},~\citeyear{qi_stanza_2020}).

The fine-tuning data were adjusted for gender-neutral wording in two rounds. Firstly, we used our own list of extracted affixed words combined with~\citeposs{vanmassenhove_neutral_2021} list to replace sexist with gender-inclusive terms. Their list covers additional word pairs like \textit{stewardess}\textendash\textit{flight attendant} or \textit{waitress}\textendash\textit{server}. 
Words that were part of named entities were not replaced.
Secondly, feminine and masculine singular pronouns (\textit{he}, \textit{she}, \textit{himself}, etc.) were re-written into the respective variants of singular \textit{they} using~\citeposs{vanmassenhove_neutral_2021} NeuTral Rewriter. 
Table \ref{tab:rewriting_example} illustrates this re-writing process and provides an example sentence within the different variants of the corpus: normal, with replacements, and rewritten with replacements.

We then reduced the final dataset, because fine-tuning a model with the entire 250 million word corpus would have gone beyond computational resources available to us and good fine-tuning results can be achieved with considerably less data~\citep{thakur-etal-2023-language, zhou_lima_2023}. We first reduced the \textit{Heap} corpus to a smaller dataset of 50 million tokens (the \textit{Small Heap}, \textasciitilde300MB), and finally only extracted lines containing word replacements. The composition of the final dataset, \textit{Tiny Heap}, can be seen in Table \ref{tab:corpus_composition}.

\subsection{Models and Fine-tuning}\label{ssec:models}
We ran our experiments on three models: GPT-2~\citep{radford_language_2019}, RoBERTa-large~\citep{liu_roberta_2019} and PHI-1.5~\citep{li_textbooks_2023}. 
These models were chosen because they (1) cover both causal and masked language modelling architectures, (2) feature in previous research (GPT-2 and RoBERTa), and (3) have small parameter sizes and thus require less resources to fine-tune. Microsoft's PHI-1.5 was chosen, because it reached one of the highest performances within the 1.5 billion parameter category of pre-trained models in Huggingface's OpenLeaderboard\footnote{\url{https://huggingface.co/spaces/HuggingFaceH4/open_llm_leaderboard}} at the time we conducted our experiments.

The models were fine-tuned for each one and three epochs (batch size 2) on an NVIDIA A100-SXM4-40GB GPU on Google Colaboratory, using 30 GPU hours in total for all models. The two fine-tuning datasets used were \textit{Tiny Heap} with gender-neutral replacements (\texttt{tiny-heap-rep}) and gender-neutral replacements and rewriting with \citeposs{vanmassenhove_neutral_2021} NeuTral Rewriter (\texttt{tiny-heap-rep-neutral}). The learning rate was set to $2\mathrm{e}{-5}$ with a weight decay of 0.01. We used the \texttt{Trainer} class of the Huggingface \texttt{transformers} library in python (version 4.38.0.dev0;~\citeauthor{wolf_transformers_2020},~\citeyear{wolf_transformers_2020}) and kept all other hyperparameters at their default values.

\subsection{Bias Evaluation Metrics}\label{ssec:bias_measure}
We utilise three established metrics for quantifying bias.
CrowS-Pairs~\citep{nangia_crows-pairs_2020} and RedditBias~\citep{barikeri_redditbias_2021} were selected because they are not based on artificial templates but are crowdsourced and extracted from naturally occurring data, respectively. The third benchmark, HONEST~\citep{nozza_honest_2021,nozza-etal-2022-measuring}, was selected as an extrinsic metric because it relies on prompt completion. In addition to measuring bias along the binary male-female axis, both RedditBias and HONEST support gender bias evaluation in relation to LGBTQ+ (Lesbian, Gay, Binary, Trans and Queer or Questioning) terminology. 

\paragraph{CrowS-Pairs}~\citep{nangia_crows-pairs_2020} is a benchmark comprised of crowdsourced minimal sentence pairs differing in words that are related to a variety of social categories, such as race, ability and gender. Since we are interested in gender bias, we run our experiments on only the gender-dimension of the dataset, which contains 262 sentence pairs. 
Each sentence pair consists of a stereotypical setting (\textit{Women don't know how to drive.}), and an anti-stereotypical setting (\textit{Men know how to drive.}), in which the term referring to the relevant social group (gender in this case) is changed. 
We use~\citeposs{meade_empirical_2022} implementation of the sentence scoring, which measures the likelihood of the changed, instead of the unchanged, tokens within a sentence. 
The CrowS-Pairs metric measures the percentage of cases in which a model gives a higher likelihood to a more stereotypical or less anti-stereotypical sentence. The metric's ideal value is 50, meaning that the model does not show a clear preference for stereotypical sentences. 

\paragraph{RedditBias}~\citep{barikeri_redditbias_2021} also contains minimal sentence pairs expressing stereotypes for different demographic dimensions: \textit{religion}, \textit{race}, \textit{gender} and \textit{queerness}. Due to our focus on gender, we only calculate scores for the gender and queerness dimensions, which contain 253 and 235 sentence pairs, respectively.
The sentences in RedditBias were extracted from the Reddit social network forum and contain both a target term identifying a social demographic (boldface in Example \ref{ex:reddit}) as well as an attribute term that expresses a (negative) stereotype related to the group (underlined in Example \ref{ex:reddit}). 
\eenumsentence{
\item \textit{his (\textbf{mom} | \textbf{dad}) is seriously devoted to \underline{cleaning}.}
\item \textit{those (\textbf{gays} | \textbf{straights}) are so \underline{dramatic}. 
}}\label{ex:reddit}

Stereotyping in a model is quantified through calculating the perplexity of the model for the sentence pairs and performing the student's t-test on the perplexity pairs. Negative values of $t$ indicate stereotypical bias in the model while $p$ indicates statistical significance of the perplexity differences. 

\paragraph{HONEST} differs from the first two measures in that it does not measure gender stereotyping but the presence of hurtful language in LLM sentence completions. The original HONEST benchmark consists of prompts containing binary masculine and feminine terms (\citealp{nozza_honest_2021}; Example \ref{ex:honest_binary}). This was later extended with prompts containing LGBTQ+ terms (\citealp{nozza-etal-2022-measuring}; Example \ref{ex:honest_queer}).
\eenumsentence{
\item \textit{the \textbf{sister} should work as a} \underline{\hspace{1cm}}.\label{ex:honest_binary}
\item \textit{The \textbf{transwoman} was hired as a} \underline{\hspace{1cm}}.\label{ex:honest_queer}	
}
HONEST uses the HurtLex lexicon of harmful language~\citep{bassignana_hurtlex_2018} to measure the hurtfulness of words contained sentence completions. HurtLex provides a classification of hurtful language into nine categories such as \textsc{animals} or \textsc{derogatory words}. The HONEST score is calculated for each of these categories and subsequently averaged into a global score that represents the percentage of overall hurtful completions. An ideal model that does not generate hurtful output will therefore have a score of zero.
For our experiments, we used $k=20$ random sentence completions for GPT-2 and RoBERTa, keeping in line with the original paper, and $k=5$ completions for PHI-1.5 in order to shorten the runs. 

\section{Results and Discussion}\label{sec:results}

\begin{table*}[t]
    \centering
    \begin{tblr}{
    colspec = {X[1.2,c]X[0.4,c]X[1.6,c]X[c]X[c]X[c]X[c]X[c]X[c]X[c]},
    rows = {m},
    row{1} = {font=\bfseries},
    column{3} = {font=\ttfamily, l},
    column{1} = {font=\bfseries},
    stretch = 0,
    hline{1,3,18} = {solid},
    row{3,8,13} = {baselinegrey},
    }
    \SetCell[r=2]{c,m}model & \SetCell[r=2]{c,m}epochs & \SetCell[r=2]{c,m}\textbf{FT} & \SetCell[c=2]{c,m} RedditBias & & \SetCell[c=3]{c,m} CrowsPairs (in\%) & & & \SetCell[c=2]{c,m} HONEST & \\
    & & & t\textsubscript{gender} & t\textsubscript{queerness} & metric & stereo & anti-st. & binary & queer \\
    \SetCell[r=5]{c,m}GPT-2 & 0 & baseline & -1.28 & -1.65 & 56.87 & 53.46 & 62.14 & 0.140 & 0.146 \\
     & \SetCell[r=2]{c,m}1 & replacement & -2.01* & \textbf{-0.39} & 54.96 & 51.57 & 60.19 & \textbf{0.101} & \textbf{0.112} \\
     & & rep+neutral & \textbf{-0.77} & -0.69 & 54.96 & 58.94 & \textbf{49.51} & 0.107 & 0.119 \\ 
     & \SetCell[r=2]{c,m} 3 & replacement & -1.54 & -0.81 & 54.58 & \textbf{49.69} & 62.14 & 0.110 & 0.120 \\
     &  & rep+neutral & -1.54 & -1.09 & \textbf{54.2} & 56.60 & \textbf{50.49} & 0.124 & 0.126 \\ \hline \hline
    \SetCell[r=5]{c,m}PHI-1.5 & 0 & baseline & \textbf{-1.83} & \textbf{-0.34} & 55.73 & 62.26 & 45.63 & \textbf{0.079} & 0.142 \\
     & \SetCell[r=2]{c,m}1 & replacement & -2.06* & -2.32* & 51.15 & \textbf{51.57} & 50.49 & 0.109 & \textbf{0.114} \\
     &  & rep+neutral & -2.26* & -2.42* & \textbf{50.76} & 55.35 & \textbf{43.69} & 0.123 & 0.154\\ 
     & \SetCell[r=2]{c,m}3 & replacement & -2.72* & -2.87* & 51.91 & 53.46 & 49.51 & 0.084 & 0.135 \\
     & 3 & rep+neutral & -2.71* & -2.16 & 51.91 & 55.97 & 45.63 & 0.093 & 0.129 \\ \hline \hline
    \SetCell[r=5]{l,m}{RoBERTa} & 0 & baseline & -0.50 & 1.50 & 60.15 & 72.15 & 42.16 & 0.035 & 0.05 \\
     & \SetCell[r=2]{c,m}1 & replacement & -0.56 & 1.42 & 50.19 & 58.23 & \textbf{38.24} & 0.044 & 0.066 \\
     &  & rep+neutral & -2.62* & -0.06 & 56.32 & 62.26 & 46.06 & 0.040 & 0.054 \\
     & \SetCell[r=2]{c,m}3 & replacement & -1.61 & 0.47 & 52.87 & 60.38 & 41.18 & \textbf{0.012} & \textbf{0.035} \\
     &  & rep+neutral & \textbf{0.22} & \textbf{2.18}* & \textbf{49.04} & \textbf{54.72} & 40.20 & 0.028 & 0.041 \\
    \end{tblr}
    \caption{Gender-stereotyping (RedditBias, CrowsPairs) and hurtful language generation (HONEST) results for different interventions to fine-tuning (FT) data, divided by baseline model, one, and three epochs of fine-tuning; RedditBias results marked * significant with $p<0.05$.
    \texttt{rep+neutral} = gender-neutral replacements + neutral rewriting; anti-st = anti-stereotypical setting}
    \label{tab:results}
\end{table*}

\subsection{Gender-marking affixes}
Table \ref{tab:extraction_rounds} illustrates the number of affixed word extractions for three rounds of verification. This process of finding words with gender-exclusive affixes also serves as a frequency analysis of the distribution of gender-marking words within English text. 
Overall, it can be clearly seen in Table \ref{tab:extraction_rounds} that gender-marking through suffixation is more common than prefixation.
Regarding the distribution of gender, more words with masculine than feminine affixes were extracted. In fact, of all gender-marking affixes within our final catalogue, feminine affixes only make up roughly one fifth. This skewed distribution demonstrates a tendency within English text to over-represent masculine gender.
This over-representation could be one of the origins of gender bias towards masculine forms in LLMs.  
Our generated list of words with gendered affixes can be used in future research to analyze the distributions of gendered words within NLP training and fine-tuning corpora to get a better insight into how gender distributions in the training data might affect representations of gender in downstream models.

\subsection{Fine-tuning}

Table \ref{tab:results} shows how fine-tuning impacted three different bias metrics for the three LLMs we tested. Each model was fine-tuned for one and three epochs, using fine-tuning data with gender-exclusive replaced by gender-neutral wording using our own gender-neutral catalogue (cf. Section \ref{ssec:extraction}) as well as~\citeposs{vanmassenhove_neutral_2021} list (\texttt{replacement}).
In addition, gender-neutral rewriting~\citep{vanmassenhove_neutral_2021} was performed on the fine-tuning data (\texttt{rep+neutral}). 

For \textbf{RedditBias}~\citep{barikeri_redditbias_2021}, we report the values of the $t$-statistic for the Student's t-test. Negative values indicate higher perplexity of the model for sentence variants mentioning female/queer target terms, which indicates stereotypical bias in the model.  
The results illustrated in Table \ref{tab:results} show \textbf{binary gender bias} for all baseline LLMs in the binary gender setting. This bias can be reduced (increasing values of $t$) by fine-tuning in the case of GPT-2 and RoBERTa. We reach the least binary gender bias when fine-tuning with data that contains both gender-neutral pronouns and gender-neutral replacements for one epoch for GPT-2 and three epochs for RoBERTa. Fine-tuning PHI-1.5 achieves opposite results, increasing the binary bias metric. 

Measuring \textbf{queerness bias}, GPT-2 exhibits the most stereotypical bias, followed by PHI-1.5, which shows a low negative value of $t_{queerness}$, indicating that the model might not be as biased towards LGBTQ+ terms as GPT-2. Even further, baseline RoBERTa shows a positive value for $t_{queerness}$ (1.5). Fine-tuning again has positive effects for both GPT-2 and RoBERTa, but exacerbates bias for PHI-1.5. Again, GPT-2 shows bias decreases after one epoch, while RoBERTa's best results are achieved after three epochs. 

For \textbf{CrowS-Pairs}~\citep{nangia_crows-pairs_2020}, we report the percentage of cases in which a model assigns higher likelihood to gendered target terms within a sentence expressing a stereotype (`stereo' column in Table \ref{tab:results}) or a lower probability to target terms in sentences expressing an anti-stereotype (`anti-st.' column in Table \ref{tab:results}). 
The `metric' column contains the overall stereotype score.
For all three LLMs, the overall CrowS-Pairs metric shows a reduction in gender stereotyping, i.e. results that are lower than the baseline and approach a value of 50\%. 
This result is mostly in line or goes beyond of what~\citet{thakur-etal-2023-language} reported for their methods of fine-tuning with gender-inclusive text; they showed a maximum reduction of the CrowS-Pairs score of approximately 2.7\% for RoBERTa-base.
Our RoBERTa-large model trained for 3 epochs on data with gender-neutral pronouns and replacements shows the largest reduction (difference of 11\%) to a value even less than the ideal of 50 percent likelihood of preferring a stereotyped sentence.
GPT-2 shows the best result (54.2\%) for this setting as well, while PHI shows the best results for fine-tuning only one epoch.
Moreover, for GPT-2 there is a tendency for fine-tuning in the \texttt{replacement} setting to lower the stereotype score, while the \texttt{replacement+neutral} setting lowers the anti-stereotype score.

The \textbf{HONEST} scores contain the percentage of sentence completions for sentences containing a term referring to binary or queer gender were completed with hurtful language. 
The two baseline causal LLMs GPT-2 and PHI-1.5 generate hurtful sentence completions around 15\% of the time in the queer setting, while RoBERTa has a much lower starting point with only 5\% hurtful completions. 
Table \ref{tab:results} shows that our method of fine-tuning language models can be used to reduce the number of hurtful completions. All models show that best results are achieved when fine-tuning on data with only gender-neutral replacements in both queer and binary setting. However, depending on the model and the setting (binary vs. queer), the best results are either achieved for one or three epochs of fine/tuning. Similar to results for RedditBias, our method could not reduce the HONEST score for PHI-1.5 in the binary setting.

\textbf{Overall}, our results echo those of ~\citet{aribandi_how_2021} who found that bias metrics within the NLP literature often do not correlate. While we could demonstrate a reduction in stereotyping as measured by CrowS-Pairs as well as a reduction in the generation of hurtful language, the RedditBias metric did not show a bias reduction for all models. 
Moreover, the fact that different models proved to be susceptible to bias reduction in different settings, such as level of gender-neutralisation in fine-tuning data or number of fine-tuning epochs, additionally shows that model specifications such as architecture and model size need to be taken into account when choosing a bias mitigation strategy. For instance, RoBERTa generally shows a larger bias reduction when fine-tuning for three epochs, while the best number of epochs for PHI-1.5 and GPT-2 depends on the fine-tuning data.
Furthermore, we demonstrated that a newer model, PHI-1.5~\citep{li_textbooks_2023}, which was released in 2023 as opposed to RoBERTa~\citep{liu_roberta_2019} and GPT-2~\citep{radford_language_2019} in 2019, was less susceptible to gender bias reduction through fine-tuning. However, the baseline PHI-1.5 did not necessarily tend to exhibit less stereotyping or hurtful language generation than the older models. 

\section{Conclusion}

Gender-inclusive language has a long history of development and advocacy within the field of feminist linguistics, but it has only recently entered gender bias research in NLP. 
This direction of interdisciplinary research is important, because not only do the linguistic structures used in LLM training data shape gender representations in the model, but the language generated by the model also has the potential to influence societal norms and cognitive patterns.
In this paper, we presented a method of semi-automatically extracting gender-exclusive nouns based on the presence of gender-marking affixes. We then extended this list with gender-neutral variants, presenting a catalogue of 692 gender-exclusive vs. -inclusive pairs, which we make available for future research. 

We further performed fine-tuning experiments on three LLMs. 
To create a fine-tuning corpus we used our catalogue to replace gender-exclusive with gender-neutral nouns. We also re-wrote gendered pronouns with the respective variants of singular \textit{they}.
Fine-tuning with gender-neutral data showed an overall reduction in gender stereotyping as measured by likelihood of gendered word generation in stereotyped settings, as well as a reduction in the generation of harmful language when prompted with sentences containing words related to binary gender as well as the LGBTQ+ community. However, we also showed that optimal bias reduction is dependent on model architecture and number of fine-tuning epochs, which need to be considered in deployment.
We hope that our work will inspire further research into the effects of gender-inclusive terminology within large language models.

\section{Limitations}

This study is limited by four main factors: 

Firstly, our study is \textbf{limited to English} specifically. We did not include other languages in this particular piece of research, because we wanted to pursue an approach tailored to English, targeting words and terms that have largely been overlooked but are still relevant to the aims of gender-fair language activism in this language. Therefore, the resources we developed and utilised, i.e. our catalogue of term-pairs, the \textit{Tiny Heap} corpus, and~\citeposs{vanmassenhove_neutral_2021} \textit{NeuTral Rewriter}, are monolingual.
Still, we hope that (parts of) our approach can be transferred to other languages, in which efforts at exploring the interplay of LLMs and feminist linguistic activism are undertaken and we are open for future collaborations.

Secondly, we performed \textbf{naive replacements} within our fine-tuning data: words found in our catalogue of gendered words were replaced with gender-neutral variants without regard for the sentence context. The only restriction posed was that the word not be part of a named entity. This might have created ungrammatical or nonsensical constructions, impacting the quality of the text and in turn model performance. Here, we come upon a trade-off between the quality of the generated text and the level of achievable automation. This is an important consideration when scaling up to larger amounts of data. Additionally, gender-exclusive terms were only replaced by a single neutral term; however, for some words several variations are possible, such as \textit{chairperson} or \textit{chair} for \textit{chairman/-woman}. Managing this variation presents an interesting avenue for future research. 

Thirdly, there is an increasing number of \textbf{bias metrics} to measure gender bias, and a growing body of work critiquing them~\citep{goldfarb-tarrant_this_2023,orgad_choose_2022
}. For example,~\citet{blodgett_stereotyping_2021} found several pitfalls in the CrowS-Pairs benchmark~\citep{nangia_crows-pairs_2020}, which we used in this paper. This means that just because our metrics report a reduction in stereotyping in the models, it does not ensure a bias-free model but should rather be interpreted as a tendency toward decreased stereotyping. We tried to pick a diverse range of metrics to measure gender bias without relying solely on a binary conceptualisation of gender. However, our choice of metrics was also limited by ease of use and interpretation. Besides issues with the bias metrics themselves, future work could additionally explore whether our fine-tuning approach impacts the performance of the models on NLU tasks. 

Lastly, our study was limited to \textbf{language models of relatively small size}. The largest models we used (GPT-2 and PHI-1.5) each have 1.5 billion parameters, which is significantly smaller than for example the smallest (seven billion parameter) model in the Llama suite of LLMs~\citep{touvron_llama_2023}, which reaches state-of-the-art performance using an open-source approach. We already demonstrated that the benefits of our approach differ based on the model used, which is why it would be interesting to see how fine-tuning with gender-neutral data impacts state-of-the-art models. 
However, our research institute does not have the resources to perform a study with models of state-of-the-art scale at the level of detail we provided here. Therefore, we leave experimentation with larger models to future research. 

\section*{Acknowledgements}
We acknowledge the Research IT HPC Service at University College Dublin for providing computational facilities and support that contributed to the research results reported in this paper.
This publication has emanated from research conducted with the financial support of Science Foundation Ireland under Grant number 12/RC/2289\_P2. For the purpose of Open Access, the authors have applied a CC BY public copyright licence to any Author Accepted Manuscript version arising from this submission.

\bibliography{anthology,zotero}

\appendix

\section{Appendix}
\label{sec:appendix}

\begin{table*}[]
    \begin{tblr}{
    colspec={X[c]X[0.3,c]X[c]X[0.3,c]X[c]X[0.3,c]X[c]X[0.3,c]},
    row{1} = {font=\bfseries},
    stretch = 0,
    }
    \hline
    \textit{man-} & \# & \textit{woman-} & \# & \textit{boy-} & \# & \textit{girl-} & \#\\ \hline
    man-made & 181 & womankind & 45 & boyfriend & 5,333 & girlfriend & 7,442 \\
    man-child & 24 & womanism & 12 & boyish & 32 & girlish & 20 \\
    man-eating & 17 & womanist & 9 & boyband & 13 & girliness & 17 \\
    man-eater & 11 & womanly & 2 & boyscout & 3 & girlfight & 5 \\
    man-crush & 10 & & & boyism & 3 & girllove & 4 \\
    man power & 10 & & & boyishly & 1 & girldom & 2 \\
    man-boobs & 9 & & & boytoy & 1 & girlification & 2 \\
    man-hater & 9 & & & & & girlfag & 1 \\
    man-hating & 7 & & & & & girlishly & 1 \\
    manstopper & 7 & & & & & girlpower & 1 \\
    \hline
    \end{tblr}
    \caption{Top 10 words with gender-denoting prefixes after second round of verification and their frequencies within 200-million token subset of OpenWebText2; empty rows indicate that $<10$ instances were found.}
    \label{tab:top_10_prefixes}
\end{table*}

\begin{table}[]
    \begin{tblr}{
    colspec={X[1.5,c]X[c]},
    row{1,22} = {font=\bfseries},
    stretch = 0,
    }
    \hline
    \textit{-manship} & \# \\ \hline
    chairmanship & 693 \\
    craftsmanship & 424 \\
    workmanship & 174 \\
    sportsmanship & 155 \\
    statesmanship & 154 \\
    showmanship & 149 \\
    marksmanship & 149 \\
    gamesmanship & 147 \\
    brinkmanship & 119 \\
    upmanship & 118 \\
    salesmanship & 105 \\
    brinksmanship & 73 \\
    penmanship & 62 \\
    seamanship & 31 \\
    swordsmanship & 28 \\
    airmanship & 21 \\
    draftsmanship & 13 \\
    horsemanship & 12 \\
    craftmanship & 6 \\
    draughtsmanship & 5 \\ \hline
    \textit{-womanship} & \# \\ \hline
    stateswomanship & 2 \\
    workwomanship & 2 \\
    \hline
    \end{tblr}
    \caption{Top 20 words with \textit{-manship} suffix and the two words with \textit{-womanship} suffix after second round of verification and their frequencies within 200-million token subset of OpenWebText2}
    \label{tab:manship_womanship}
\end{table}

\begin{table*}[]
    \begin{tblr}{
    colspec = {X},
    hlines,
    row{1,3,5,7,9,11,13,15,17,19} = {font=\bfseries, baselinegrey},
    stretch = 0,
    }
suffix: -woman \\
ambulancewoman::emergency medical technician, anchorwoman::anchorperson, anti-woman::misogynist, antiwoman::misogynist, bogeywoman::monster, bondwoman::slave, businesswoman::businessperson, cavewoman::caveperson, charwoman::cleaner, congresswoman::congressperson, craftswoman::craftsoerson, everywoman::ordinary person, fisherwoman::fisher, forewoman::foreperson, frontierswoman::explorer, frontwoman::frontperson, gentlewoman::refined person, hitwoman::assassin, horsewoman::equestrian, madwoman::maniac \\
suffix: -womanship \\
stateswomanship::statespersonship, workwomanship::workpersonship \\
suffix: -girl \\
babygirl::baby, ballgirl::ball person, bargirl::bartender, callgirl::sex worker, cavegirl::caveperson, cowgirl::cow herder, fangirl::fan, farmgirl::farm worker, papergirl::newspaper delivery person, playgirl::player, showgirl::performer, slavegirl::slave, snowgirl::snowperson, tomgirl::timid child \\
suffix: -man \\
adman::advertiser, almsman::medical social worker, ambulanceman::emergency medical technician, anchorman::anchorperson, artilleryman::cannoneer, assemblyman::assembly member, assman::assperson, backwoodsman::explorer, bagman::travelling salesperson, bargeman::barge operator, barman::bartender, baseman::baseperson, batsman::batter, bellman::bellhop, binman::garbage collector, bluesman::bluesperson, boatman::boater, bogeyman::monster, bondman::slave, bondsman::slave \\
suffix: -manship \\
airmanship::aerial skill, batsmanship::batting skill, brinkmanship::extreme strategy, brinksmanship::extreme strategy, chairmanship::chairpersonship, churchmanship ::churchpersonship, craftmanship::craftpersonship, craftsmanship::craftspersonship, draftsmanship::draftspersonship, draughtsmanship::draughtspersonship, foremanship::forepersonship, gamesmanship::unsporting tactic, gentlemanship::refinedness, grantsmanship::grant acquisition expertise, handcraftsmanship::handcraftspersonship, horsemanship::equestrian skill, journeymanship::artisanship, manship::courage, marksmanship::sharpshooting skill, oarsmanship::rowing skill \\
suffix: -boy \\
ballboy::ball person, batboy::bat person, bellboy::bellhop, busboy::restaurant attendant, callboy::sex worker, copyboy::junior newspaper worker, cowboy::cow herder, doughboy::foot soldier, fanboy::fan, farmboy::farm worker, femboy::effeminate person, fisherboy::young fisher, fratboy::fraternity member, headboy::student leader, homeboy::fellow member, houseboy::domestic worker, ladyboy::genderqueer person, nancyboy::nancy, newsboy::newspaper delivery person, paperboy::newspaper delivery person \\ \hline
prefix: woman- \\
womanism::feminism, womanist::feminist, womankind::humankind, womanly::feminine \\
prefix: girl- \\
girldom::feminine sphere, girlfag::woman attracted to gay men, girlfight::fight, girlfriend::partner, girlification::feminization, girliness::femininity, girlish::feminine, girlishly::childishly, girllove::love, girlpower::power \\
prefix: man- \\
man cave::sanctuary, man hater::hater, man hating::misandry, man hug::pound hug, man hunt::organized search, man magnet::attractive person, man marking::marking, man servant::servant, man up::adult up, man-ass::ass, man-bag::handbag, man-boobs::boobs, man-cave::sanctuary, man-cession::recession, man-child::child, man-crush::crush, man-eater::cannibal, man-eating::human-eating, man-friend::friend, man-hater::hater \\
prefix: boy- \\
boyband::band, boyfriend::partner, boyish::childish, boyishly::childishly, boyism::childism, boyscout::scout, boytoy::toy \\
    \end{tblr}
    \caption{Example terms (SG) from catalogue of gender-exclusive terms and gender-inclusive replacements; each category contains 20 example pairs or the number of pairs in the catalogue if there are $<20$ singular pairs}
    \label{tab:catalogue}
\end{table*}

\end{document}